\pdfoutput=1
\documentclass{article}

\usepackage[final,dblblindworkshop,nonatbib]{neurips_2026}
\workshoptitle{Self-Evolving Diversity-Driven Search for Robust AI Systems}
\usepackage[numbers,sort&compress]{natbib}
\usepackage[T1]{fontenc}
\usepackage{amsmath,amssymb,amsthm}
\usepackage{graphicx}
\usepackage{booktabs}
\usepackage{multirow}
\usepackage{xcolor}
\usepackage{colortbl}
\usepackage{microtype}
\usepackage{url}
\usepackage{listings}
\usepackage{algorithm}
\usepackage{algpseudocode}
\usepackage[most]{tcolorbox}
\algrenewcommand\algorithmicrequire{\textbf{Input:}}
\algrenewcommand\algorithmicensure{\textbf{Output:}}
\algrenewcommand{\algorithmiccomment}[1]{\hfill\textit{\footnotesize // #1}}
\lstdefinestyle{promptstyle}{
  basicstyle=\ttfamily\scriptsize,
  breaklines=true, breakatwhitespace=false,
  columns=fullflexible, keepspaces=true,
  frame=none, xleftmargin=0pt,
  aboveskip=0pt, belowskip=0pt,
}
\definecolor{promptsys}{RGB}{152,188,227}
\definecolor{promptsysbg}{RGB}{233,240,250}
\definecolor{promptusr}{RGB}{178,183,190}
\definecolor{promptusrbg}{RGB}{242,243,245}
\tcbset{promptboxbase/.style={
  enhanced, breakable,
  boxrule=0.6pt, arc=2pt,
  left=4pt, right=4pt, top=2pt, bottom=2pt,
  toptitle=2pt, bottomtitle=2pt,
  coltitle=black, fonttitle=\bfseries\small,
  attach boxed title to top left={yshift=-2mm, xshift=4mm},
}}
\newtcolorbox{systempromptbox}[1][]{promptboxbase,
  colback=promptsysbg, colframe=promptsys, title={#1},
  boxed title style={colback=promptsys, colframe=promptsys, arc=1.5pt, boxrule=0pt}}
\newtcolorbox{userpromptbox}[1][]{promptboxbase,
  colback=promptusrbg, colframe=promptusr, title={#1},
  boxed title style={colback=promptusr, colframe=promptusr, arc=1.5pt, boxrule=0pt}}
\usepackage{hyperref}
\definecolor{sizeblue}{RGB}{29,120,235}
\hypersetup{
    colorlinks=true,
    linkcolor=sizeblue,
    citecolor=sizeblue,
    urlcolor=sizeblue,
}
\usepackage[capitalise,noabbrev,nameinlink]{cleveref}
\crefformat{section}{\S#2#1#3}
\Crefformat{section}{\S#2#1#3}
\crefrangeformat{section}{\S\S#3#1#4--#5#2#6}
\crefmultiformat{section}{\S#2#1#3}{ and~\S#2#1#3}{, \S#2#1#3}{ and~\S#2#1#3}

\graphicspath{{figures/}}
\definecolor{tblhead}{RGB}{219,232,248}
\definecolor{tblrow}{RGB}{246,249,253}
\definecolor{tblours}{RGB}{198,220,245}

\newcommand{\tesearch}{\textsc{tesearch}}
\newcommand{\ask}{a}
\newcommand{\landed}{s}
\newcommand{\dataset}{TestGenEval}
\newtheorem{definition}{Definition}

\title{Small but Mighty: Test Suites as Improvement Signals\\
for Self-Evolving Agents}

\author{%
  Yutong Cheng$^{1}$\thanks{Work done during the internship at NEC Laboratories America.} \quad
  Wei Cheng$^{2}$ \quad
  Wenchao Yu$^{2}$ \quad
  Dawei Zhou$^{1}$ \quad
  Haifeng Chen$^{2}$ \\[2pt]
  $^{1}$Virginia Tech \qquad $^{2}$NEC Laboratories America \\[2pt]
  \texttt{\{yutongcheng,zhoud\}@vt.edu} \quad
  \texttt{\{weicheng,wyu,haifeng\}@nec-labs.com}
}

\begin{document}
\maketitle

\begin{abstract}
A self-evolving agent rewrites its own harness and must decide which rewrites to
keep, so its improvement signal is the check a rewrite has to pass. Verifying a rewrite by agent rollout is expensive and subjective, while
executable checks such as the repository's own tests are too sparse to catch
behavior-changing rewrites, rejecting only a quarter of them. We instead take test feedback as the
executable signal, and search for test suites that are both small and strong
with a quality--diversity evolutionary algorithm, so that the signal is dense
and informative.
\tesearch{} treats suite size as a behavior descriptor, so that quality is
pursued at every size and a small, strong suite is reachable, and learns a
transition kernel during the search that denoises the LLM as an actuator on
that descriptor.
Across $129$ real-world modules and three backbones,
\tesearch{} delivers the highest-quality suites of any method compared,
LLM-driven evolutionary search included, and the smallest of any LLM-based one.
Used as the improvement signal of a self-evolving agent, it rejects $59.9\%$ of
behavior-changing rewrites against $25.3\%$ for the repository's tests, at a
$4.0\%$ false-alarm rate and $3.5$\,s per decision. The code and data are
available at \url{https://github.com/ccsnow127/TeSearch}.
\end{abstract}

\section{Introduction}

Self-evolving agents get stronger by editing the code that runs them: the
harness that prompts the model, parses its actions, executes them and records
the trajectory~\citep{lou2026autoharness,wang2026harnessbridge,zhang2025dgm}.
The check a rewrite must pass to be kept is the agent's \emph{improvement
signal}. Accept a rewrite that silently changes behavior and every later episode
and rewrite inherits it; reject rewrites that preserve behavior and the
evolution stalls. In software engineering the executable form of this
check is the test suite, a \emph{regression specification}~\citep{yoo2012regression}
that captures the accepted behavior of the code so that later modification can
preserve it.

The signals available today are weak. Accepting on rollout outcome spends an
agent episode per candidate on a noisy end-to-end score.
The repository's own tests are free, but they were written before the agent
existed and anticipate none of its edits: we find them rejecting only $25.3\%$
of the rewrites that change behavior (\cref{sec:exp-selfevolve}). A generated
suite can do better, but only if it is \emph{effective}, so that a behavioral
change breaks at least one test, and \emph{efficient}, since it runs on every
proposed rewrite and its failures are read by a model with a finite context.
Existing generators, search-based or LLM-based, optimize effectiveness and
leave size to whatever
emerges~\citep{fraser2011evosuite,lukasczyk2022pynguin,alshahwan2024automated,jain2024testgeneval}.

We formulate \emph{size-optimized unit test generation} and, rather than
scalarizing quality against size, let \textbf{size define where a suite
competes and quality whether it wins}: size is a \emph{behavior descriptor} of a
quality--diversity (QD) search~\citep{mouret2015illuminating,pierrot2022multi},
so the small end stays alive instead of being eliminated by the large one. Two challenges follow. LLM calls are expensive, so a finite budget must be
\emph{allocated} across niches whose value is unknown. And unlike classical QD,
the descriptor cannot be set directly: the search can only ask the LLM for a
count, which LLMs follow imprecisely~\citep{zhou2023instruction}, so the
niche a call targets and the niche it lands in differ.

\tesearch{} answers both: an online \emph{landing kernel} models the LLM as
a noisy actuator on the descriptor, correcting its bias and predicting where
each call lands, and a \emph{size-annealing UCB} allocates calls across niches
from those predictions, turning toward smaller suites only once quality is
secured. Its suites are the strongest and, among LLM-based methods, the
smallest on $129$ real-world modules, and as the improvement signal of a
self-evolving agent they more than double the detection of behavior-changing
rewrites at a cost the loop can afford on every one.

\section{Related work}
\label{sec:related}

\textbf{Self-evolving agents and improvement signals.} Self-improving LLM
systems differ first in where an improvement is written. \emph{Weight-space}
methods fine-tune the model on its own data or
rewards~\citep{wang2023selfinstruct,zweiger2025seal,zuo2025ttrl}, a gain that
cannot be read, diffed or reverted. \emph{Context-space} methods write into the
harness instead, through verbal
self-refinement~\citep{madaan2023selfrefine,shinn2024reflexion}, prompt search
and textual gradients~\citep{yang2024llmoptimizers,yuksekgonul2024textgrad,miprov2,gepa},
experience memory~\citep{ace}, skill synthesis~\citep{wang2023voyager} or search
over the whole scaffold~\citep{hu2024adas}; at the frontier the agent rewrites
its own code, including the procedure that performs the
update~\citep{fernando2024promptbreeder,zelikman2023stop,zhang2025dgm,lou2026autoharness}.
What decides whether an improvement is kept ranges from the model's own
judgment~\citep{bai2022constitutional,yuan2024selfrewarding} to external
verifiers such as tests~\citep{chen2023codet,chen2024selfdebug},
compilers~\citep{le2022coderl,haluptzok2023teach} and
environments~\citep{ma2024eureka}. Tests are the signal coding agents such as
Cursor, Codex and Claude
Code~\citep{anysphere2026cursor,openai2025codex,anthropic2025claudecode} lean
on across whole
repositories~\citep{jimenez2024swebench,yang2024sweagent,zhang2023repocoder},
yet they are taken as given: the repository's own or written ad hoc. We
optimize the test signal itself, for the strength and cost a loop running it
on every rewrite needs.
See \cref{app:additional_related} for additional related work.

\textbf{Evolutionary and quality--diversity search.} LLMs used as variation
operators under executable fitness make evolutionary search a strong
inference-time optimizer~\citep{romera2024mathematical,alphaevolve,shinkaevolve},
but a global objective concentrates the population around one
mode~\citep{goldberg1987genetic,deb2002fast}. QD optimization instead partitions
the search space by behavior descriptors and optimizes quality within each
region~\citep{lehman2011abandoning,cully2018quality}: MAP-Elites keeps an elite
per cell~\citep{mouret2015illuminating}, and MOME a Pareto front per cell whose
hypervolumes it sums~\citep{pierrot2022multi}. Search-based test generation
evolves program inputs against coverage
goals~\citep{fraser2011evosuite,lukasczyk2022pynguin} and test-suite reduction
shrinks a written suite post hoc~\citep{harrold1993methodology,rothermel2002empirical};
neither controls size during generation. In classical QD the descriptor is read off what an
individual turns out to be; here it must be \emph{requested} from an LLM before
it can be observed, and \tesearch{} models that actuation error explicitly.

\section{Size-optimized test generation}
\label{sec:setup}

Given a Python module $M$, a \emph{suite} $T$ is a finite set of test functions
for $M$, with size $\landed(T)$. Two objectives are measured on it:
\emph{coverage} $c(T)\in[0,1]$, the fraction of executable lines of $M$
executed, and \emph{mutation score} $m(T)\in[0,1]$, the fraction of a fixed
sample of mutants of $M$ killed~\citep{jia2011analysis}. With the origin as
reference, a suite has hypervolume
$\mathrm{hv}(T)=c(T)\,m(T)$~\citep{zitzler1999multiobjective}, high only when
both objectives are.

\begin{definition}[Size-optimized unit test generation]
\label{def:task}
Given a module $M$ and executable quality metrics $(c, m)$, produce a collection
$\mathcal{T}$ of suites and deliver the smallest suite of maximal hypervolume:
\vspace{-0.4cm}

\begin{equation*}
T^\star \;=\; \operatorname*{arg\,min}_{T \in \mathcal{T}^{\max}} \landed(T),
\qquad
\mathcal{T}^{\max} \;=\; \operatorname*{arg\,max}_{T \in \mathcal{T}} \mathrm{hv}(T).
\end{equation*}
\vspace{-0.4cm}

\end{definition}

\textbf{Quality--diversity formulation.} \Cref{def:task} is the terminal
objective; QD is the search surrogate. Larger suites are
naturally advantaged in a global competition on
$(c,m)$~\citep{namin2009influence,inozemtseva2014coverage}, so size becomes the
behavior descriptor and quality is optimized independently within size niches,
each keeping a Pareto front over $(c,m)$ rather than a single
elite~\citep{pierrot2022multi}:
\vspace{-0.4cm}

\begin{equation}
\label{eq:archive}
\mathcal{A}_j = \operatorname*{nd}_{(c,m)}
\big\{\, T \in \mathcal{T} : b_{j-1} \le \landed(T) < b_j \big\},
\qquad j = 1,\dots,K,
\end{equation}
\vspace{-0.4cm}

where $\operatorname{nd}$ takes the non-dominated subset and
$b_0 < \dots < b_K$ are the niche boundaries, the outermost two left open
($b_0=0$, $b_K=\infty$). The niches are geometric, since the variation in
generated size grows with the size requested (\cref{app:grid}).

\begin{figure}[t]
  
\centering
\includegraphics[width=\textwidth]{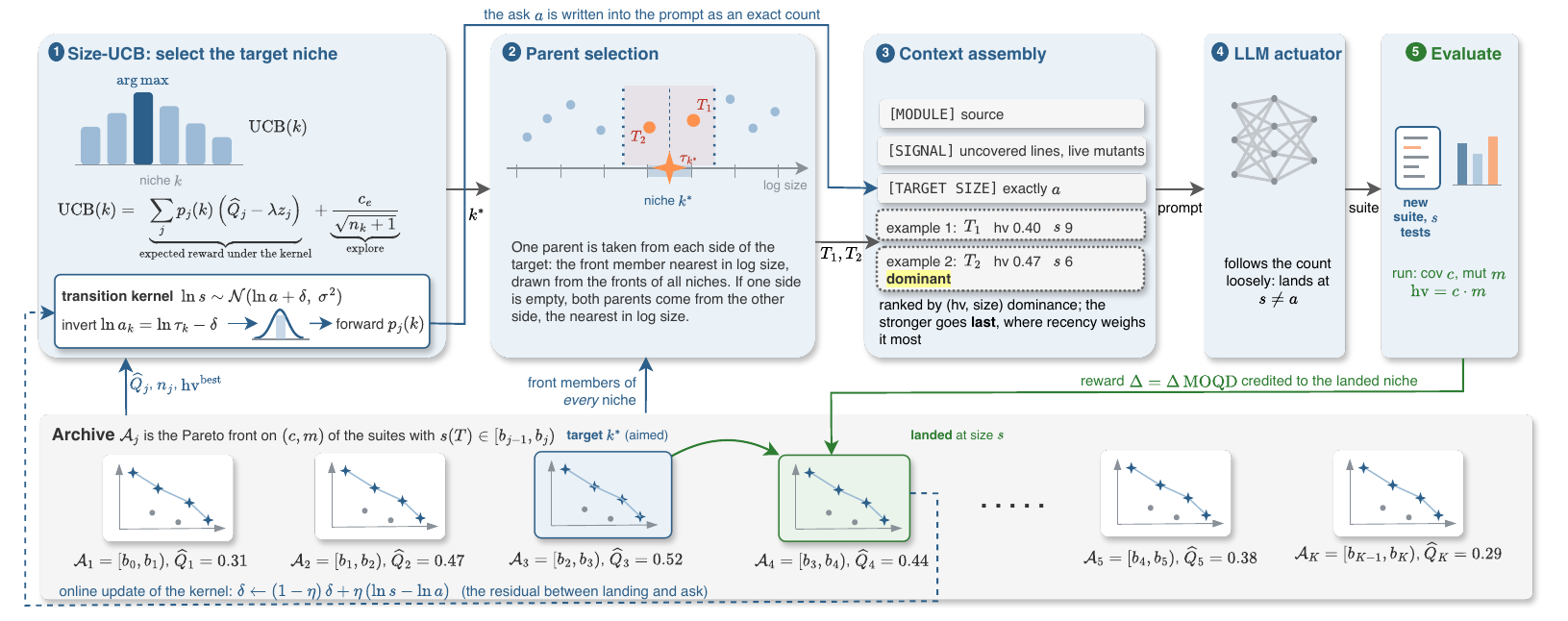}
\vspace{-0.3cm}

\caption{\textbf{Overview of \tesearch{}.} The archive (bottom) keeps a Pareto
front on (coverage, mutation score) per size niche. \textbf{(1)}~Size-UCB picks
a target niche through the transition kernel; \textbf{(2)}~the nearest front
suite on each side of the target size becomes a parent; \textbf{(3)}~parents,
module, execution feedback and target size form the context; \textbf{(4)}~the
LLM writes a suite, possibly off the requested size; \textbf{(5)}~the suite is
filed where it lands, and the size error updates the kernel.}

\label{fig:overview}
\end{figure}

\textbf{Search loop and challenges.} Each iteration (\cref{fig:overview})
selects a target niche; takes as parents the Pareto-front suite nearest the
target in log size on each side of it, ordered weaker to stronger since an LLM
weighs its most recent context most; prompts the LLM with the parents, the
module annotated with its uncovered lines and surviving mutants, and the target
size (\cref{app:parents,app:prompts}); then scores the suite and files it
where it actually lands (\cref{alg:loop}). The two
challenges are \emph{allocation} of the call budget and \emph{actuation}.

\section{\tesearch{}}
\label{sec:method}

\paragraph{The landing kernel.} Requesting a suite of size $\ask$ may produce
one of size $\landed \neq \ask$, so a call aimed at one niche may land in
another and credit assigned to the target goes to the wrong arm. LLMs follow
count instructions with multiplicative
noise~\citep{zhang2025lifebench,li2024ruler,xiao2026track}, so we model landing
in log space,
\begin{equation}
\label{eq:kernel}
\ln \landed \;\sim\; \mathcal{N}\big(\ln \ask + \delta,\ \sigma^2\big),
\end{equation}
with $\delta$ the systematic count bias and $\sigma$ the residual spread. The
bias varies across modules and backbones and enters the request directly, so it
is estimated online and compensated before generation: for a target niche $k$ with center
$\tau_k=\sqrt{b_{k-1}b_k}$, inverting \cref{eq:kernel} gives the request
$\ln \ask_k = \ln \tau_k - \delta$, rounded to an integer. The spread only
spreads credit over neighboring niches; it is stable and fixed at its holdout
median. The landing distribution
of a call aimed at niche $k$ is
\begin{equation}
\label{eq:pvec}
p_j(k) = \Phi\!\Big(\tfrac{\ln b_{j} - \mu_k}{\sigma}\Big) -
      \Phi\!\Big(\tfrac{\ln b_{j-1} - \mu_k}{\sigma}\Big), \qquad \mu_k = \ln \ask_k + \delta,
\end{equation}
with $\Phi$ the standard normal distribution function, and once the suite is
measured the bias is updated by the LMS delta
rule~\citep{widrow1960adaptive},
$\delta \leftarrow (1-\eta)\,\delta + \eta\,(\ln \landed - \ln \ask)$.
\vspace{-0.4cm}

\paragraph{Size-annealing UCB.} Niche selection is a bandit whose arms are the
niches. The reward of a call is the increment $\Delta$ it adds to the
multi-objective QD score~\citep{pierrot2022multi},
$\mathrm{MOQD}(\mathcal{A}) = \sum_j \mathrm{HV}(\mathcal{A}_j)$, the total
area the niche fronts dominate, credited to the niche the suite actually lands
in, whether it improves a niche or populates an empty one. The acquisition is
\begin{equation}
\label{eq:acq}
\mathrm{UCB}(k) = \underbrace{\sum_j p_j(k)\big(\widehat{Q}_j - \lambda z_j\big)}
                  _{\text{expected exploitation}}
                + \underbrace{\frac{c_e}{\sqrt{n_k+1}}}_{\text{exploration}},
\qquad
\lambda = \max\big(0,\ \mathrm{hv}^{\mathrm{best}} - H\big),
\end{equation}
\vspace{-0.3cm}

where $\widehat{Q}_j \leftarrow (1-\alpha)\widehat{Q}_j + \alpha\Delta$ is the
discounted reward of landing in niche $j$, initialized for an unseen niche to
the mean over observed ones; $z_j=(j-1)/(K-1)$ is the normalized niche index;
$n_k$ counts the calls aimed at niche $k$ wherever they landed; and $c_e$ is an
exploration coefficient~\citep{auer2002finite}. Two details are load-bearing.
The kernel weights the \emph{exploitation} term, because $p_j(k)$ is an
actuator outcome distribution rather than a policy
prior~\citep{auer2002finite,jones1998efficient}; and the size penalty sits
\emph{inside} the expectation, so what is penalized is the size a suite lands
at, not the size aimed for. The size penalty
$\lambda$ stays at zero until the best hypervolume found so far reaches a
threshold $H$, and grows with it afterwards. The search therefore secures
quality first and only then favors smaller suites, following the
quality-then-size order of \cref{def:task}.

\section{Experiments}
\label{sec:exp}

\subsection{Setup}
\label{sec:exp-setup}

We use the $129$ distinct Python modules of \dataset{}~\citep{jain2024testgeneval},
each with a containerized harness. Mutants are produced per module by
\texttt{cosmic-ray}~\citep{cosmicray} under a fixed seed and split into a search
fold, which scores the search, and a held-out eval fold, which scores what is
delivered. Every method runs under an identical budget of $N=50$ calls per
module, on GPT-5.4, Claude Opus 4.8 and Gemini 3.1 Pro. The baselines are
\emph{(i)}~Pynguin~\citep{lukasczyk2022pynguin}, the reference search-based tool
for Python, matched in wall clock ($6000$\,s) since it makes no model call; \emph{(ii)}~inference-time scaling: iterative
refinement~\citep{shinn2024reflexion},
best-of-$N$~\citep{li2022competition,chen2024alphamath}, beam search of width
$5$, and the context optimizers MIPROv2 and ACE~\citep{miprov2,ace};
GEPA~\citep{gepa}, whose single global Pareto pool over $(c,m,-\landed)$ makes
it the three-objective alternative to our formulation; and
OpenEvolve~\citep{openevolve}, an open implementation of
AlphaEvolve~\citep{alphaevolve} that evolves suites in a MAP-Elites archive
under the same feedback; \emph{(iii)}~the coding agents Claude Code and
Cursor~\citep{anthropic2025claudecode,anysphere2026cursor}, told to maximize
quality and minimize size, with harness feedback as tool output; and \emph{(iv)}~ablations: \emph{w/o kernel} asks for
the niche center as written and credits each arm where its calls land;
\emph{w/o UCB} sweeps the grid uniformly while keeping the aim; \emph{w/o
nearest parents} samples parents at random from the fronts; \emph{w/o
$\lambda$} sets $\lambda=0$.

\begin{table}[t]
\caption{Delivered suite on the $129$ \dataset{} modules, means over modules,
same call budget for every method. Higher is better except \textbf{size}, ranked
separately; the smallest per backbone is in \textcolor{sizeblue}{blue}. Claude Code runs only on its own
backbone. $^*$Pynguin is matched in wall clock and scored on the $40$ modules it
supports.}
\vspace{-0.2cm}
\label{tab:main}
\scriptsize
\setlength{\tabcolsep}{3.4pt}
\renewcommand{\arraystretch}{1.12}
\resizebox{\textwidth}{!}{%
\begin{tabular}{ll|cccc|cccc|cccc}
\toprule
\rowcolor{tblhead}
 & Method & cov & mut & hv & size & \multicolumn{8}{c}{} \\
\midrule
SBST & Pynguin$^*$ & 0.464 & 0.157 & 0.090 & 20.4
 &  \\
\midrule
\rowcolor{tblhead}
 & & \multicolumn{4}{c|}{GPT-5.4} & \multicolumn{4}{c|}{Claude Opus 4.8}
       & \multicolumn{4}{c}{Gemini 3.1 Pro} \\
\rowcolor{tblhead}
\multirow{-2}{*}{} &  & cov & mut & hv & size & cov & mut & hv & size
 & cov & mut & hv & size \\
\midrule
\multirow{7}{*}{\shortstack[l]{Inference-time\\scaling}} & Iter.\ refinement & 0.798 & 0.647 & 0.516 & 50.8 & 0.842 & 0.849 & 0.716 & 72.5 & 0.694 & 0.764 & 0.539 & 56.8 \\
 & Best-of-$N$ & 0.908 & 0.758 & 0.693 & 47.6 & 0.953 & 0.861 & 0.824 & 61.5 & 0.837 & 0.823 & 0.691 & 51.4 \\
 & Beam search & 0.859 & 0.905 & 0.775 & 44.5 & 0.971 & 0.910 & 0.886 & 68.8 & 0.890 & 0.934 & 0.838 & 52.6 \\
 & MIPROv2 & 0.927 & 0.802 & 0.746 & 48.3 & 0.972 & 0.891 & 0.870 & 63.1 & 0.859 & 0.853 & 0.736 & 53.1 \\
 & GEPA & 0.929 & 0.786 & 0.732 & 47.9 & 0.982 & 0.911 & 0.895 & 67.2 & 0.845 & 0.857 & 0.732 & 54.7 \\
 & ACE & 0.914 & 0.753 & 0.692 & 52.4 & 0.938 & 0.878 & 0.827 & 64.4 & 0.835 & 0.826 & 0.692 & 55.5 \\
 & OpenEvolve & 0.963 & 0.858 & 0.831 & 51.5 & 0.964 & 0.917 & 0.885 & 68.0 & 0.930 & 0.853 & 0.798 & 56.6 \\
\midrule
\multirow{2}{*}{\shortstack[l]{Coding\\agents}} & Claude Code & --- & --- & --- & --- & 0.982 & 0.908 & 0.891 & 51.6 & --- & --- & --- & --- \\
 & Cursor & 0.913 & 0.762 & 0.699 & 44.4 & 0.967 & 0.878 & 0.852 & 48.2 & 0.846 & 0.826 & 0.701 & 37.0 \\
\midrule
\multirow{4}{*}{Ablations} & w/o kernel & 0.974 & 0.884 & 0.866 & 40.1 & 0.988 & 0.929 & 0.917 & 40.8 & 0.905 & 0.942 & 0.855 & 63.3 \\
 & w/o UCB & 0.973 & 0.881 & 0.861 & 44.1 & 0.995 & 0.964 & 0.960 & 34.4 & 0.906 & 0.937 & 0.851 & 42.3 \\
 & w/o near.\ parents & 0.985 & 0.857 & 0.847 & 39.2 & 0.984 & 0.949 & 0.934 & 33.2 & 0.922 & 0.937 & 0.864 & 39.4 \\
 & w/o $\lambda$ & 0.981 & 0.880 & 0.865 & 58.8 & 0.991 & 0.961 & 0.954 & 84.6 & 0.909 & 0.934 & 0.852 & 73.4 \\
\midrule
\rowcolor{tblours} & \textbf{\tesearch{} (full)} & \textbf{0.995} & \textbf{0.920} & \textbf{0.916} & \textcolor{sizeblue}{\textbf{38.6}} & \textbf{0.998} & \textbf{0.977} & \textbf{0.976} & \textcolor{sizeblue}{\textbf{29.1}} & \textbf{0.940} & \textbf{0.963} & \textbf{0.908} & \textcolor{sizeblue}{\textbf{36.3}} \\
\bottomrule
\end{tabular}}
\vspace{-0.2cm}
\end{table}

\subsection{Test-suite quality}
\label{sec:exp-main}

\Cref{tab:main} reports the delivered suite. \tesearch{} attains the highest
hypervolume on every backbone, $7.0$--$8.5$ points above the strongest
baseline, in the smallest suites any LLM-based method delivers. Pynguin delivers
fewer tests, but at $\mathrm{hv}=0.090$: size is a coordinate here, not the
objective. The two population-based baselines isolate the formulation.
OpenEvolve also keeps an archive and is the strongest baseline on GPT-5.4, but
its archive is not indexed by size, so it holds nothing small to hand back and
delivers $51.5$ tests. GEPA puts $-\landed$ in the objective, admitting near-empty
suites as non-dominated and leaving a front wide in size but shallow in
quality. The ablations separate on the axis
each mechanism moves: removing $\lambda$ costs little quality ($-0.02$ to
$-0.06$ hv) but grows suites $1.5$--$2.9\times$; removing the kernel inflates
size hardest where the actuator overshoots most ($+11.7$ and $+27.0$ tests on
Claude Opus 4.8 and Gemini 3.1 Pro), since credit piles up in the larger niches
that off-target calls reach; removing UCB or the nearest parents costs quality
and size on all three backbones.

\textbf{The actuator and the anneal.} Over $6063$ \tesearch{} generations on
GPT-5.4, the count bias $\delta$ ranges from $-0.037$ to $+0.107$ across modules
and changes sign, so no fixed correction serves all; online, the log-size
residuals have mean $0.013$ and standard deviation $0.100$, as the kernel
assumes. Allocation is selective without collapsing: grouped into eight
normalized size bands, no band falls below a $0.067$ share of calls in any
window. Against constant and no size pressure, only the quality-gated ramp
finishes both highest in quality and smallest in size.

\subsection{Improvement signal for a self-evolving agent}
\label{sec:exp-selfevolve}

\textbf{Setup.} We study the keep-or-discard decision on
mini-swe-agent~\citep{minisweagent}, an open variant of
SWE-agent~\citep{yang2024sweagent} whose harness is a few hundred lines of
Python. For each of eight harness modules, GPT-5.4 receives a revision intent
drawn from common self-improvement goals (latency, tokens, robustness,
structure), either behavior-preserving or behavior-changing, and implements it
as a revised module, $64$ per module; each behavior-changing intent is the
failure mode of a preserving one with the same goal and shape of edit, such as
caching a function that reads changing state (\cref{app:revisions}). Two annotators verify each
revision ($\kappa=0.74$), leaving the $470$ of
\cref{tab:signal}, labeled \emph{divergent} if the intent changes behavior and
\emph{equivalent} otherwise. A gate rejects a revision when its suite fails
where it passed on the unmodified module; the \tesearch{} suite is generated
once from that module and compared with its repository tests.

\begin{table}[t]
\caption{Improvement signals for self-revisions of mini-swe-agent's harness.
div/eq: divergent / equivalent revisions in the module's pool. DDR: divergent
revisions rejected ($\uparrow$); FAR: equivalent revisions rejected
($\downarrow$); s: mean wall clock per decision.}

\label{tab:signal}
\centering
\footnotesize
\setlength{\tabcolsep}{4.5pt}
\renewcommand{\arraystretch}{1.05}
\resizebox{\textwidth}{!}{%
\begin{tabular}{ll|c|ccc|ccc}
\toprule
\rowcolor{tblhead}
 & & & \multicolumn{3}{c|}{Repository tests} & \multicolumn{3}{c}{\tesearch{}} \\
\rowcolor{tblhead}
Module & Role in the harness & div/eq & DDR & FAR & s & DDR & FAR & s \\
\midrule
\texttt{agents/default} & agent loop: model calls, actions, limits & 17/41 & 0.529 & 0.000 & 15.7 & \textbf{0.588} & 0.049 & 3.0 \\
\texttt{environments/local} & runs bash commands, captures output & 25/31 & 0.720 & 0.000 & 4.9 & \textbf{0.920} & 0.065 & 3.3 \\
\texttt{actions\_toolcall} & parses tool calls; renders observations & 29/34 & 0.276 & 0.000 & 3.0 & \textbf{0.552} & 0.029 & 3.1 \\
\texttt{actions\_text} & parses text actions; renders observations & 25/35 & 0.040 & 0.000 & 7.7 & \textbf{0.560} & 0.029 & 2.9 \\
\texttt{cache\_control} & places prompt-cache breakpoints & 29/26 & 0.172 & 0.000 & 2.9 & \textbf{0.414} & 0.038 & 3.0 \\
\texttt{content\_string} & flattens messages to text for display & 35/25 & 0.057 & 0.000 & 2.9 & \textbf{0.600} & 0.040 & 2.9 \\
\texttt{litellm\_model} & model client: API call, retry, cost & 38/25 & 0.026 & 0.000 & 7.1 & \textbf{0.421} & 0.080 & 6.7 \\
\texttt{serialize} & merges layered config dictionaries & 19/36 & 0.579 & 0.000 & 2.9 & \textbf{0.947} & 0.000 & 3.1 \\
\midrule
\rowcolor{tblours}
Pooled &  & 217/253 & 0.253 & 0.000 & 5.9 & \textbf{0.599} & 0.040 & 3.5 \\
\bottomrule
\end{tabular}}
\vspace{-0.4cm}
\end{table}

\textbf{Results.} \tesearch{} rejects $59.9\%$ of divergent revisions against
$25.3\%$ for the repository tests, and more on every module
(\cref{tab:signal}). Repository tests check that a module performs its function;
a mutation-optimized suite also pins the details a mutant perturbs, such as a
separator, a default or which exception surfaces, so the gap is widest on
modules that parse model outputs or render what the model reads. The tighter constraint costs little: \tesearch{} rejects $10$ of the
$253$ equivalent revisions, a $4.0\%$ false-alarm rate, where the repository
tests reject none. A decision takes $3.5$\,s against $5.9$\,s and calls no
model, where acceptance by rollout spends an agent episode: the agent can
afford this signal on every rewrite it proposes.

\section{Conclusion}

Self-evolving agents now rewrite their own harness fluently; what they cannot
yet tell is whether a self-referential edit is correct and leaves existing
behavior intact. This paper targets that gap with a quality--diversity
evolutionary search that treats suite size as a behavior descriptor, so quality
is pursued at every size and a small, strong suite is reachable, and learns a transition kernel during the search that denoises the LLM as an actuator on that descriptor. Across $129$ modules and three backbones the
resulting suites are the strongest of any method compared and the smallest of
any LLM-based one, and as the gate of a self-evolving agent they reject
$59.9\%$ of behavior-changing rewrites against $25.3\%$ for the repository's
tests, at $3.5$\,s per decision and no model call. \tesearch{} is thus an
effective improvement signal for self-evolving agents: strong enough to catch
what a rewrite silently broke, cheap enough to run on every rewrite.

\bibliographystyle{plainnat-nourl}
\bibliography{evorobust}

\begin{thebibliography}{74}
\providecommand{\natexlab}[1]{#1}
\providecommand{\url}[1]{\texttt{#1}}
\expandafter\ifx\csname urlstyle\endcsname\relax
  \providecommand{\doi}[1]{doi: #1}\else
  \providecommand{\doi}{doi: \begingroup \urlstyle{rm}\Url}\fi

\bibitem[Agrawal et~al.(2026)Agrawal, Tan, Soylu, Ziems, Khare, Opsahl-Ong,
  Singhvi, Shandilya, Ryan, Jiang, et~al.]{gepa}
Lakshya~A. Agrawal, Shangyin Tan, Dilara Soylu, Noah Ziems, Rishi Khare, Krista
  Opsahl-Ong, Arnav Singhvi, Herumb Shandilya, Michael~J. Ryan, Meng Jiang,
  et~al.
\newblock {GEPA}: Reflective prompt evolution can outperform reinforcement
  learning.
\newblock In \emph{ICLR}, 2026.

\bibitem[Alshahwan et~al.(2024)Alshahwan, Chheda, Finogenova, Gokkaya, Harman,
  Harper, Marginean, Sengupta, and Wang]{alshahwan2024automated}
Nadia Alshahwan, Jubin Chheda, Anastasia Finogenova, Beliz Gokkaya, Mark
  Harman, Inna Harper, Alexandru Marginean, Shubho Sengupta, and Eddy Wang.
\newblock Automated unit test improvement using large language models at
  {M}eta.
\newblock In \emph{FSE Companion}, 2024.

\bibitem[{Anthropic}(2025)]{anthropic2025claudecode}
{Anthropic}.
\newblock Claude {C}ode: an agentic coding tool, 2025.
\newblock \url{https://claude.com/claude-code}.

\bibitem[{Anysphere}(2026)]{anysphere2026cursor}
{Anysphere}.
\newblock {Cursor}: {AI} coding agent for building ambitious software, 2026.
\newblock \url{https://cursor.com}.

\bibitem[Auer et~al.(2002)Auer, Cesa-Bianchi, and Fischer]{auer2002finite}
Peter Auer, Nicol{\`o} Cesa-Bianchi, and Paul Fischer.
\newblock Finite-time analysis of the multiarmed bandit problem.
\newblock \emph{Machine Learning}, 2002.

\bibitem[Bai et~al.(2022)Bai, Kadavath, Kundu, Askell,
  et~al.]{bai2022constitutional}
Yuntao Bai, Saurav Kadavath, Sandipan Kundu, Amanda Askell, et~al.
\newblock Constitutional {AI}: Harmlessness from {AI} feedback.
\newblock \emph{arXiv preprint arXiv:2212.08073}, 2022.

\bibitem[Chen et~al.(2023)Chen, Zhang, Nguyen, Zan, Lin, Lou, and
  Chen]{chen2023codet}
Bei Chen, Fengji Zhang, Anh Nguyen, Daoguang Zan, Zeqi Lin, Jian-Guang Lou, and
  Weizhu Chen.
\newblock {CodeT}: Code generation with generated tests.
\newblock In \emph{ICLR}, 2023.

\bibitem[Chen et~al.(2024{\natexlab{a}})Chen, Liao, Li, and
  Fan]{chen2024alphamath}
Guoxin Chen, Minpeng Liao, Chengxi Li, and Kai Fan.
\newblock {AlphaMath} almost zero: Process supervision without process.
\newblock In \emph{NeurIPS}, 2024{\natexlab{a}}.

\bibitem[Chen et~al.(2024{\natexlab{b}})Chen, Lin, Sch{\"a}rli, and
  Zhou]{chen2024selfdebug}
Xinyun Chen, Maxwell Lin, Nathanael Sch{\"a}rli, and Denny Zhou.
\newblock Teaching large language models to self-debug.
\newblock In \emph{ICLR}, 2024{\natexlab{b}}.

\bibitem[Cheng et~al.(2025)Cheng, Bajaber, Tsegai, Song, and
  Gao]{cheng2025ctinexus}
Yutong Cheng, Osama Bajaber, Saimon~Amanuel Tsegai, Dawn Song, and Peng Gao.
\newblock {CTINexus}: Automatic cyber threat intelligence knowledge graph
  construction using large language models.
\newblock In \emph{IEEE European Symposium on Security and Privacy (EuroS\&P)},
  2025.

\bibitem[Cheng et~al.(2026{\natexlab{a}})Cheng, Chen, Gao, and
  Cheng]{cheng2026specforge}
Yutong Cheng, Haifeng Chen, Peng Gao, and Wei Cheng.
\newblock {SpecForge}: Recursive self-improvement for coding agents via
  behavioral specifications.
\newblock In \emph{NeurIPS}, 2026{\natexlab{a}}.

\bibitem[Cheng et~al.(2026{\natexlab{b}})Cheng, Chen, Yu, Zhao, Gao, and
  Cheng]{cheng2026docsearch}
Yutong Cheng, Haifeng Chen, Wenchao Yu, Xujiang Zhao, Peng Gao, and Wei Cheng.
\newblock Escaping whack-a-mole: Optimizing documentation as repo-specific
  playbooks for coding agents.
\newblock In \emph{ICML}, 2026{\natexlab{b}}.

\bibitem[Cheng et~al.(2026{\natexlab{c}})Cheng, Li, Cui, Ding, Wang, Chen, and
  Gao]{cheng2026ctifoundry}
Yutong Cheng, Changze Li, Qian Cui, Wei Ding, Lingzhi Wang, Yan Chen, and Peng
  Gao.
\newblock {CTIFoundry}: An agent-native corpus scaffold for cyber threat
  intelligence.
\newblock \emph{arXiv preprint arXiv:2608.18613}, 2026{\natexlab{c}}.

\bibitem[Cheng et~al.(2026{\natexlab{d}})Cheng, Liu, Li, Song, and
  Gao]{cheng2026cticonnect}
Yutong Cheng, Yang Liu, Changze Li, Dawn Song, and Peng Gao.
\newblock {CTIConnect}: A benchmark for retrieval-augmented {LLMs} over
  heterogeneous cyber threat intelligence.
\newblock In \emph{KDD}, 2026{\natexlab{d}}.

\bibitem[Cully and Demiris(2018)]{cully2018quality}
Antoine Cully and Yiannis Demiris.
\newblock Quality and diversity optimization: A unifying modular framework.
\newblock \emph{IEEE Transactions on Evolutionary Computation}, 2018.

\bibitem[Deb et~al.(2002)Deb, Pratap, Agarwal, and Meyarivan]{deb2002fast}
Kalyanmoy Deb, Amrit Pratap, Sameer Agarwal, and T.~Meyarivan.
\newblock A fast and elitist multiobjective genetic algorithm: {NSGA-II}.
\newblock \emph{IEEE Transactions on Evolutionary Computation}, 2002.

\bibitem[Fernando et~al.(2024)Fernando, Banarse, Michalewski, Osindero, and
  Rockt{\"a}schel]{fernando2024promptbreeder}
Chrisantha Fernando, Dylan Banarse, Henryk Michalewski, Simon Osindero, and Tim
  Rockt{\"a}schel.
\newblock Promptbreeder: Self-referential self-improvement via prompt
  evolution.
\newblock In \emph{ICML}, 2024.

\bibitem[Fraser and Arcuri(2011)]{fraser2011evosuite}
Gordon Fraser and Andrea Arcuri.
\newblock {EvoSuite}: automatic test suite generation for object-oriented
  software.
\newblock In \emph{ESEC/FSE}, 2011.

\bibitem[Goldberg and Richardson(1987)]{goldberg1987genetic}
David~E. Goldberg and Jon Richardson.
\newblock Genetic algorithms with sharing for multimodal function optimization.
\newblock In \emph{ICGA}, 1987.

\bibitem[Haluptzok et~al.(2023)Haluptzok, Bowers, and
  Kalai]{haluptzok2023teach}
Patrick Haluptzok, Matthew Bowers, and Adam~Tauman Kalai.
\newblock Language models can teach themselves to program better.
\newblock In \emph{ICLR}, 2023.

\bibitem[Harrold et~al.(1993)Harrold, Gupta, and Soffa]{harrold1993methodology}
Mary~Jean Harrold, Rajiv Gupta, and Mary~Lou Soffa.
\newblock A methodology for controlling the size of a test suite.
\newblock \emph{ACM Transactions on Software Engineering and Methodology},
  1993.

\bibitem[Hu et~al.(2025)Hu, Lu, and Clune]{hu2024adas}
Shengran Hu, Cong Lu, and Jeff Clune.
\newblock Automated design of agentic systems.
\newblock In \emph{ICLR}, 2025.

\bibitem[Inozemtseva and Holmes(2014)]{inozemtseva2014coverage}
Laura Inozemtseva and Reid Holmes.
\newblock Coverage is not strongly correlated with test suite effectiveness.
\newblock In \emph{ICSE}, 2014.

\bibitem[Jain et~al.(2025)Jain, Synnaeve, and Rozi{\`e}re]{jain2024testgeneval}
Kush Jain, Gabriel Synnaeve, and Baptiste Rozi{\`e}re.
\newblock {TestGenEval}: A real world unit test generation and test completion
  benchmark.
\newblock In \emph{ICLR}, 2025.

\bibitem[Jia and Harman(2011)]{jia2011analysis}
Yue Jia and Mark Harman.
\newblock An analysis and survey of the development of mutation testing.
\newblock \emph{IEEE Transactions on Software Engineering}, 2011.

\bibitem[Jimenez et~al.(2024)Jimenez, Yang, Wettig, Yao, Pei, Press, and
  Narasimhan]{jimenez2024swebench}
Carlos~E. Jimenez, John Yang, Alexander Wettig, Shunyu Yao, Kexin Pei, Ofir
  Press, and Karthik Narasimhan.
\newblock {SWE}-bench: Can language models resolve real-world {GitHub} issues?
\newblock \emph{International Conference on Learning Representations}, 2024.

\bibitem[Jones et~al.(1998)Jones, Schonlau, and Welch]{jones1998efficient}
Donald~R. Jones, Matthias Schonlau, and William~J. Welch.
\newblock Efficient global optimization of expensive black-box functions.
\newblock \emph{Journal of Global Optimization}, 13\penalty0 (4):\penalty0
  455--492, 1998.

\bibitem[Lange et~al.(2026)Lange, Imajuku, and Cetin]{shinkaevolve}
Robert~Tjarko Lange, Yuki Imajuku, and Edoardo Cetin.
\newblock {ShinkaEvolve}: Towards open-ended and sample-efficient program
  evolution.
\newblock In \emph{ICLR}, 2026.

\bibitem[Le et~al.(2022)Le, Wang, Gotmare, Savarese, and Hoi]{le2022coderl}
Hung Le, Yue Wang, Akhilesh~Deepak Gotmare, Silvio Savarese, and Steven C.~H.
  Hoi.
\newblock {CodeRL}: Mastering code generation through pretrained models and
  deep reinforcement learning.
\newblock In \emph{NeurIPS}, 2022.

\bibitem[Lehman and Stanley(2011)]{lehman2011abandoning}
Joel Lehman and Kenneth~O. Stanley.
\newblock Abandoning objectives: Evolution through the search for novelty
  alone.
\newblock \emph{Evolutionary Computation}, 2011.

\bibitem[Li et~al.(2024)Li, Zhang, Li, Liu, Bai, Luo, Chen, and
  Yang]{li2024ruler}
Jiaming Li, Lei Zhang, Yunshui Li, Ziqiang Liu, Yuelin Bai, Run Luo, Longze
  Chen, and Min Yang.
\newblock Ruler: A model-agnostic method to control generated length for large
  language models.
\newblock In \emph{Findings of EMNLP}, 2024.

\bibitem[Li et~al.(2022)Li, Choi, Chung, Kushman, Schrittwieser, Leblond,
  Eccles, Keeling, Gimeno, Dal~Lago, et~al.]{li2022competition}
Yujia Li, David Choi, Junyoung Chung, Nate Kushman, Julian Schrittwieser,
  R{\'e}mi Leblond, Tom Eccles, James Keeling, Felix Gimeno, Agustin Dal~Lago,
  et~al.
\newblock Competition-level code generation with {AlphaCode}.
\newblock \emph{Science}, 2022.

\bibitem[Liu et~al.(2024)Liu, Lin, Hewitt, Paranjape, Bevilacqua, Petroni, and
  Liang]{liu2024lost}
Nelson~F. Liu, Kevin Lin, John Hewitt, Ashwin Paranjape, Michele Bevilacqua,
  Fabio Petroni, and Percy Liang.
\newblock Lost in the middle: How language models use long contexts.
\newblock \emph{Transactions of the Association for Computational Linguistics},
  2024.

\bibitem[Lou et~al.(2026)Lou, L{\'a}zaro-Gredilla, Dedieu, Wendelken, Lehrach,
  and Murphy]{lou2026autoharness}
Xinghua Lou, Miguel L{\'a}zaro-Gredilla, Antoine Dedieu, Carter Wendelken,
  Wolfgang Lehrach, and Kevin~P. Murphy.
\newblock {AutoHarness}: improving {LLM} agents by automatically synthesizing a
  code harness.
\newblock \emph{arXiv preprint arXiv:2603.03329}, 2026.

\bibitem[Lukasczyk and Fraser(2022)]{lukasczyk2022pynguin}
Stephan Lukasczyk and Gordon Fraser.
\newblock Pynguin: Automated unit test generation for {P}ython.
\newblock In \emph{ICSE Companion}, 2022.

\bibitem[Ma et~al.(2024)Ma, Liang, Wang, Huang, Bastani, Jayaraman, Zhu, Fan,
  and Anandkumar]{ma2024eureka}
Yecheng~Jason Ma, William Liang, Guanzhi Wang, De-An Huang, Osbert Bastani,
  Dinesh Jayaraman, Yuke Zhu, Linxi Fan, and Anima Anandkumar.
\newblock Eureka: Human-level reward design via coding large language models.
\newblock In \emph{ICLR}, 2024.

\bibitem[Madaan et~al.(2023)Madaan, Tandon, Gupta, Hallinan, Gao, Wiegreffe,
  Alon, Dziri, Prabhumoye, Yang, Gupta, Majumder, Hermann, Welleck,
  Yazdanbakhsh, and Clark]{madaan2023selfrefine}
Aman Madaan, Niket Tandon, Prakhar Gupta, Skyler Hallinan, Luyu Gao, Sarah
  Wiegreffe, Uri Alon, Nouha Dziri, Shrimai Prabhumoye, Yiming Yang, Shashank
  Gupta, Bodhisattwa~Prasad Majumder, Katherine Hermann, Sean Welleck, Amir
  Yazdanbakhsh, and Peter Clark.
\newblock Self-refine: Iterative refinement with self-feedback.
\newblock In \emph{NeurIPS}, 2023.

\bibitem[Masson et~al.(2019)Masson, Rim, and Lee]{masson2019ddsketch}
Charles Masson, Jee~E. Rim, and Homin~K. Lee.
\newblock {DDSketch}: A fast and fully-mergeable quantile sketch with
  relative-error guarantees.
\newblock \emph{Proceedings of the VLDB Endowment}, 2019.

\bibitem[Mouret and Clune(2015)]{mouret2015illuminating}
Jean-Baptiste Mouret and Jeff Clune.
\newblock Illuminating search spaces by mapping elites.
\newblock \emph{arXiv preprint arXiv:1504.04909}, 2015.

\bibitem[Namin and Andrews(2009)]{namin2009influence}
Akbar~Siami Namin and James~H. Andrews.
\newblock The influence of size and coverage on test suite effectiveness.
\newblock In \emph{ISSTA}, 2009.

\bibitem[Novikov et~al.(2025)Novikov, V{\~u}, Eisenberger, Dupont, Huang,
  Wagner, Shirobokov, Kozlovskii, Ruiz, Mehrabian, et~al.]{alphaevolve}
Alexander Novikov, Ng{\^a}n V{\~u}, Marvin Eisenberger, Emilien Dupont, Po-Sen
  Huang, Adam~Zsolt Wagner, Sergey Shirobokov, Borislav Kozlovskii, Francisco
  J.~R. Ruiz, Abbas Mehrabian, et~al.
\newblock {AlphaEvolve}: A coding agent for scientific and algorithmic
  discovery.
\newblock \emph{arXiv preprint arXiv:2506.13131}, 2025.

\bibitem[{OpenAI}(2025)]{openai2025codex}
{OpenAI}.
\newblock Introducing {Codex}, 2025.
\newblock \url{https://openai.com/index/introducing-codex/}.

\bibitem[{OpenEvolve contributors}(2025)]{openevolve}
{OpenEvolve contributors}.
\newblock {OpenEvolve}: An open-source implementation of {AlphaEvolve}, 2025.
\newblock \url{https://github.com/algorithmicsuperintelligence/openevolve}.

\bibitem[Opsahl-Ong et~al.(2024)Opsahl-Ong, Ryan, Purtell, Broman, Potts,
  Zaharia, and Khattab]{miprov2}
Krista Opsahl-Ong, Michael~J. Ryan, Josh Purtell, David Broman, Christopher
  Potts, Matei Zaharia, and Omar Khattab.
\newblock Optimizing instructions and demonstrations for multi-stage language
  model programs.
\newblock In \emph{EMNLP}, 2024.

\bibitem[Pierrot et~al.(2022)Pierrot, Richard, Beguir, and
  Cully]{pierrot2022multi}
Thomas Pierrot, Guillaume Richard, Karim Beguir, and Antoine Cully.
\newblock Multi-objective quality diversity optimization.
\newblock In \emph{GECCO}, 2022.

\bibitem[Putra et~al.(2026)Putra, Basuki, Cheng, and Gao]{putra2026nl2logic}
Rizky~Ramadhana Putra, Raihan Sultan~Pasha Basuki, Yutong Cheng, and Peng Gao.
\newblock {NL2Logic}: {AST} guided translation of natural language into
  first-order logic with large language models.
\newblock In \emph{Findings of EACL}, 2026.

\bibitem[Romera-Paredes et~al.(2024)Romera-Paredes, Barekatain, Novikov, Balog,
  Kumar, Dupont, Ruiz, Ellenberg, Wang, Fawzi, Kohli, and
  Fawzi]{romera2024mathematical}
Bernardino Romera-Paredes, Mohammadamin Barekatain, Alexander Novikov, Matej
  Balog, M.~Pawan Kumar, Emilien Dupont, Francisco J.~R. Ruiz, Jordan~S.
  Ellenberg, Pengming Wang, Omar Fawzi, Pushmeet Kohli, and Alhussein Fawzi.
\newblock Mathematical discoveries from program search with large language
  models.
\newblock \emph{Nature}, 2024.

\bibitem[Rothermel et~al.(2002)Rothermel, Harrold, von Ronne, and
  Hong]{rothermel2002empirical}
Gregg Rothermel, Mary~Jean Harrold, Jeffery von Ronne, and Christie Hong.
\newblock Empirical studies of test-suite reduction.
\newblock \emph{Software Testing, Verification and Reliability}, 2002.

\bibitem[Shinn et~al.(2023)Shinn, Cassano, Gopinath, Narasimhan, and
  Yao]{shinn2024reflexion}
Noah Shinn, Federico Cassano, Ashwin Gopinath, Karthik Narasimhan, and Shunyu
  Yao.
\newblock Reflexion: Language agents with verbal reinforcement learning.
\newblock In \emph{NeurIPS}, 2023.

\bibitem[{Sixty North}(2015)]{cosmicray}
{Sixty North}.
\newblock Cosmic ray: mutation testing for {Python}, 2015.
\newblock \url{https://github.com/sixty-north/cosmic-ray}.

\bibitem[{SWE-agent team}(2025)]{minisweagent}
{SWE-agent team}.
\newblock mini-{SWE}-agent: the 100-line {AI} agent that solves {GitHub}
  issues, 2025.
\newblock \url{https://github.com/SWE-agent/mini-swe-agent}.

\bibitem[van Deursen et~al.(2001)van Deursen, Moonen, van~den Bergh, and
  Kok]{vandeursen2001refactoring}
Arie van Deursen, Leon Moonen, Alex van~den Bergh, and Gerard Kok.
\newblock Refactoring test code.
\newblock Technical Report SEN-R0119, CWI, 2001.

\bibitem[van Erven and Koolen(2016)]{vanerven2016metagrad}
Tim van Erven and Wouter~M. Koolen.
\newblock Metagrad: Multiple learning rates in online learning.
\newblock In \emph{NeurIPS}, 2016.

\bibitem[Wang et~al.(2023{\natexlab{a}})Wang, Xie, Jiang, Mandlekar, Xiao, Zhu,
  Fan, and Anandkumar]{wang2023voyager}
Guanzhi Wang, Yuqi Xie, Yunfan Jiang, Ajay Mandlekar, Chaowei Xiao, Yuke Zhu,
  Linxi Fan, and Anima Anandkumar.
\newblock Voyager: An open-ended embodied agent with large language models.
\newblock \emph{arXiv preprint arXiv:2305.16291}, 2023{\natexlab{a}}.

\bibitem[Wang et~al.(2026)Wang, Wang, Taylor, Cong, Sun, and
  Wang]{wang2026harnessbridge}
Xiaoxuan Wang, Haixin Wang, Alexander Taylor, Jason Cong, Yizhou Sun, and Wei
  Wang.
\newblock {HarnessBridge}: learnable bidirectional controller for {LLM} agent
  harness.
\newblock \emph{arXiv preprint arXiv:2606.12882}, 2026.

\bibitem[Wang et~al.(2023{\natexlab{b}})Wang, Kordi, Mishra, Liu, Smith,
  Khashabi, and Hajishirzi]{wang2023selfinstruct}
Yizhong Wang, Yeganeh Kordi, Swaroop Mishra, Alisa Liu, Noah~A. Smith, Daniel
  Khashabi, and Hannaneh Hajishirzi.
\newblock Self-instruct: Aligning language models with self-generated
  instructions.
\newblock In \emph{ACL}, 2023{\natexlab{b}}.

\bibitem[Widrow and Hoff(1988)]{widrow1960adaptive}
Bernard Widrow and Marcian~E. Hoff.
\newblock Adaptive switching circuits.
\newblock In \emph{Neurocomputing: Foundations of Research}. MIT Press, 1988.

\bibitem[Xiao et~al.(2026)Xiao, Wang, Hu, Miao, Shen, Wang, Luo, and
  Su]{xiao2026track}
Meiman Xiao, Ante Wang, Qingguo Hu, Zhongjian Miao, Huangjun Shen, Longyue
  Wang, Weihua Luo, and Jinsong Su.
\newblock Can {LLMs} track their output length? a dynamic feedback mechanism
  for precise length regulation.
\newblock \emph{arXiv preprint arXiv:2601.01768}, 2026.

\bibitem[Xu et~al.(2023)Xu, Fu, Zhu, Cheng, Li, Ma, and He]{xu2023hue}
Junjielong Xu, Qiuai Fu, Zhouruixing Zhu, Yutong Cheng, Zhijing Li, Yuchi Ma,
  and Pinjia He.
\newblock Hue: A user-adaptive parser for hybrid logs.
\newblock In \emph{ESEC/FSE}, 2023.

\bibitem[Yang et~al.(2024{\natexlab{a}})Yang, Wang, Lu, Liu, Le, Zhou, and
  Chen]{yang2024llmoptimizers}
Chengrun Yang, Xuezhi Wang, Yifeng Lu, Hanxiao Liu, Quoc~V. Le, Denny Zhou, and
  Xinyun Chen.
\newblock Large language models as optimizers.
\newblock In \emph{ICLR}, 2024{\natexlab{a}}.

\bibitem[Yang et~al.(2024{\natexlab{b}})Yang, Jimenez, Wettig, Lieret, Yao,
  Narasimhan, and Press]{yang2024sweagent}
John Yang, Carlos~E. Jimenez, Alexander Wettig, Kilian Lieret, Shunyu Yao,
  Karthik Narasimhan, and Ofir Press.
\newblock {SWE}-agent: Agent-computer interfaces enable automated software
  engineering.
\newblock In \emph{NeurIPS}, 2024{\natexlab{b}}.

\bibitem[Yoo and Harman(2012)]{yoo2012regression}
Shin Yoo and Mark Harman.
\newblock Regression testing minimization, selection and prioritization: a
  survey.
\newblock \emph{Software Testing, Verification and Reliability}, 2012.

\bibitem[Yuan et~al.(2024)Yuan, Pang, Cho, Li, Sukhbaatar, Xu, and
  Weston]{yuan2024selfrewarding}
Weizhe Yuan, Richard~Yuanzhe Pang, Kyunghyun Cho, Xian Li, Sainbayar
  Sukhbaatar, Jing Xu, and Jason Weston.
\newblock Self-rewarding language models.
\newblock In \emph{ICML}, 2024.

\bibitem[Yuksekgonul et~al.(2024)Yuksekgonul, Bianchi, Boen, Liu, Huang,
  Guestrin, and Zou]{yuksekgonul2024textgrad}
Mert Yuksekgonul, Federico Bianchi, Joseph Boen, Sheng Liu, Zhi Huang, Carlos
  Guestrin, and James Zou.
\newblock {TextGrad}: Automatic ``differentiation'' via text.
\newblock \emph{arXiv preprint arXiv:2406.07496}, 2024.

\bibitem[Zelikman et~al.(2024)Zelikman, Lorch, Mackey, and
  Kalai]{zelikman2023stop}
Eric Zelikman, Eliana Lorch, Lester Mackey, and Adam~Tauman Kalai.
\newblock Self-taught optimizer ({STOP}): Recursively self-improving code
  generation.
\newblock In \emph{COLM}, 2024.

\bibitem[Zhang et~al.(2023)Zhang, Chen, Zhang, Keung, Liu, Zan, Mao, Lou, and
  Chen]{zhang2023repocoder}
Fengji Zhang, Bei Chen, Yue Zhang, Jacky Keung, Jin Liu, Daoguang Zan, Yi~Mao,
  Jian-Guang Lou, and Weizhu Chen.
\newblock {RepoCoder}: Repository-level code completion through iterative
  retrieval and generation.
\newblock In \emph{EMNLP}, 2023.

\bibitem[Zhang et~al.(2026{\natexlab{a}})Zhang, Hu, Lu, Lange, and
  Clune]{zhang2025dgm}
Jenny Zhang, Shengran Hu, Cong Lu, Robert Lange, and Jeff Clune.
\newblock Darwin {G}{\"o}del machine: Open-ended evolution of self-improving
  agents.
\newblock In \emph{ICLR}, 2026{\natexlab{a}}.

\bibitem[Zhang et~al.(2026{\natexlab{b}})Zhang, Hu, Upasani, Ma, Hong,
  Kamanuru, et~al.]{ace}
Qizheng Zhang, Changran Hu, Shubhangi Upasani, Boyuan Ma, Fenglu Hong,
  Vamsidhar Kamanuru, et~al.
\newblock Agentic context engineering: Evolving contexts for self-improving
  language models.
\newblock In \emph{ICLR}, 2026{\natexlab{b}}.

\bibitem[Zhang et~al.(2025)Zhang, Zhou, Wang, Fang, Xu, Zhang, Wang, Zhang, Li,
  Sun, Lyu, Liu, and Su]{zhang2025lifebench}
Wei Zhang, Zhenhong Zhou, Kun Wang, Junfeng Fang, Rongwu Xu, Yuanhe Zhang, Rui
  Wang, Ge~Zhang, Xinfeng Li, Li~Sun, Lingjuan Lyu, Yang Liu, and Sen Su.
\newblock {LIFEBench}: Evaluating length instruction following in large
  language models.
\newblock In \emph{NeurIPS Datasets and Benchmarks Track}, 2025.

\bibitem[Zhao et~al.(2021)Zhao, Wallace, Feng, Klein, and
  Singh]{zhao2021calibrate}
Zihao Zhao, Eric Wallace, Shi Feng, Dan Klein, and Sameer Singh.
\newblock Calibrate before use: Improving few-shot performance of language
  models.
\newblock In \emph{ICML}, 2021.

\bibitem[Zhou et~al.(2023)Zhou, Lu, Mishra, Brahma, Basu, Luan, Zhou, and
  Hou]{zhou2023instruction}
Jeffrey Zhou, Tianjian Lu, Swaroop Mishra, Siddhartha Brahma, Sujoy Basu,
  Yi~Luan, Denny Zhou, and Le~Hou.
\newblock Instruction-following evaluation for large language models.
\newblock \emph{arXiv preprint arXiv:2311.07911}, 2023.

\bibitem[Zitzler and Thiele(1999)]{zitzler1999multiobjective}
Eckart Zitzler and Lothar Thiele.
\newblock Multiobjective evolutionary algorithms: a comparative case study and
  the strength {P}areto approach.
\newblock \emph{IEEE Transactions on Evolutionary Computation}, 1999.

\bibitem[Zuo et~al.(2025)Zuo, Zhang, Sheng, Qu, Cui, Zhu, Li, Zhang, Long, Hua,
  Qi, Sun, Ma, Yuan, Ding, and Zhou]{zuo2025ttrl}
Yuxin Zuo, Kaiyan Zhang, Li~Sheng, Shang Qu, Ganqu Cui, Xuekai Zhu, Haozhan Li,
  Yuchen Zhang, Xinwei Long, Ermo Hua, Biqing Qi, Youbang Sun, Zhiyuan Ma,
  Lifan Yuan, Ning Ding, and Bowen Zhou.
\newblock {TTRL}: Test-time reinforcement learning.
\newblock In \emph{NeurIPS}, 2025.

\bibitem[Zweiger et~al.(2025)Zweiger, Pari, Guo, Aky{\"u}rek, Kim, and
  Agrawal]{zweiger2025seal}
Adam Zweiger, Jyothish Pari, Han Guo, Ekin Aky{\"u}rek, Yoon Kim, and Pulkit
  Agrawal.
\newblock Self-adapting language models.
\newblock In \emph{NeurIPS}, 2025.

\end{thebibliography}

\appendix

\section{The size-niched grid}
\label{app:grid}
\label{app:design}

Suite size is the archive's only descriptor, and the noise on it is
multiplicative: asking for twice as many tests does not add a fixed error, it
scales the error, which is why \cref{sec:method} models landing as log-normal.
The grid is therefore a partition of $\ln \landed$ built once, immediately after
the seed phase. The grid is centered on an \emph{anchor} $\bar{s}$, the size the
backbone writes at when no size is named, estimated as the geometric mean of the
seed suites' sizes. Writing $T_1, \dots, T_n$ for the $n$ seed suites that
measured,
{\setlength{\abovedisplayskip}{5pt}\setlength{\belowdisplayskip}{5pt}
\begin{equation}
\label{eq:anchor}
\bar{s} \;=\; \exp\!\Big(\tfrac{1}{n}\sum\nolimits_{i=1}^{n} \ln \landed(T_i)\Big).
\end{equation}}%
The mean is taken on the log scale because the noise is multiplicative, and the
seeds are unhinted precisely so that the anchor measures the model rather than
our own prompt.

Niches are geometric, with equal width on the log scale. This is the spacing
used whenever values within a constant factor of each other are
interchangeable, as in MetaGrad's exponentially spaced grid of learning rates
\citep{vanerven2016metagrad}, and whenever error is relative rather than
absolute, as in DDSketch, whose bucket ratio is set by the relative accuracy it
guarantees \citep{masson2019ddsketch}. Under multiplicative noise, both hold for
suite sizes. Consecutive niches differ by the ratio $\rho = e^{2\sigma} = 1.22$,
where $\sigma = 0.10$ is the landing spread of the transition kernel
(\cref{sec:method}), so each niche spans $\pm\sigma$ around its center on the
log scale. This width is intended: since \cref{sec:method} models landing noise
as log-normal, an aimed draw lands in the niche it was aimed at with probability
$68\%$ under the model.

The grid is built per module over the range $[s_{\min},\, \bar{s}\rho^{\gamma}]$, and
both ends follow conventions of software testing
\citep{vandeursen2001refactoring,namin2009influence}. The lower end is set
by the module: $s_{\min}$ is the number of public functions and methods of the
module under test, counted statically from its AST. A suite with fewer tests
than that must check several of them in one test method, which is the
\emph{Eager Test} smell: such a test is hard to read, hard to use as
documentation of the behavior it checks, and makes tests more dependent on each
other, and the prescribed remedy is to split it until each test method exercises
a single method of the unit \citep{vandeursen2001refactoring}. The floor also
guards against reward hacking, in which the LLM packs all its checks into a
single test function to game the size reward. The upper end is relative to the
anchor, $\gamma$ niches above it, and so scales with the size of the module.
In software testing, the effectiveness of a suite grows with its size
\citep{inozemtseva2014coverage}, but only logarithmically
\citep{namin2009influence}. Beyond a minimal suite, one from which no test can be
removed without lowering its mutation score, further tests add nothing but code
smells such as \emph{Lazy Test} and \emph{Test Code Duplication}
\citep{vandeursen2001refactoring}. To keep suite size from growing without bound,
and to avoid spending calls on unreasonably large suites, we set
$\gamma = 4$. The boundaries that fall inside the range, sorted
and with duplicates dropped, are $b_1 < \dots < b_{K-1}$. A suite outside the
range is filed rather than dropped, into one of two open overflow niches,
$[0, b_1)$ and $[b_{K-1}, \infty)$; the bandit never aims at them, and a call
that lands in one is still charged to the arm it aimed at.

\section{The search loop}
\label{app:alg}

\Cref{alg:loop} states the whole method.

\begin{algorithm}[t]
\caption{\tesearch{} on one module. $\textsc{llm}$ is the frozen backbone;
$\textsc{measure}$ executes a suite in the module's container and returns
$(c, m)$ on its passing subset together with its written size $\landed$.}
\label{alg:loop}
\begin{algorithmic}[1]
\Require module $M$; call budget $N$; seeds $n_{\mathrm{seed}}$; hyperparameters
$\sigma, \eta, \alpha, c_e, H$
\Ensure one delivered suite
\Statex \emph{Seed phase --- the grid does not exist yet, so nothing can be discarded on it.}
\For{$i = 1, \dots, n_{\mathrm{seed}}$}
  \State $T_i \gets \textsc{llm}(\textsc{seed}, M)$
    \Comment{unhinted: no count is named}
  \State $\textsc{measure}(T_i)$; file $T_i$ provisionally, dominance off
\EndFor
\State $\bar{s} \gets$ Eq.~\eqref{eq:anchor} on the measured seed sizes;\ \
       $s_{\min} \gets$ public functions and methods of $M$
\State derive $b_0 \dots b_K$ over $[s_{\min},\, \bar{s}\rho^{\gamma}]$
       (\cref{app:grid}); re-file the archive under them
\State $\delta \gets 0$;\ \ $n_k \gets 0$ for all $k$;\ \
       $\mathrm{hv}^{\mathrm{best}} \gets \max_i \mathrm{hv}(T_i)$
\State replay the seed landings into $\widehat{Q}$
    \Comment{a seed asked for no size}
\Statex \emph{Search.}
\While{fewer than $N$ calls have been spent}
  \If{the archive is empty}
    \Comment{only after every seed failed to measure}
    \State regenerate unhinted, measure, file; \textbf{continue}
  \EndIf
  \For{$k = 2, \dots, K-1$}
    \Comment{the overflow niches $1$ and $K$ are never aimed at}
    \State $P_k \gets$ the front member nearest $\tau_k$ in log size on each
           side of $\tau_k$ \Comment{\cref{app:parents}}
    \State $\ask_k \gets \big\lfloor \exp(\ln \tau_k - \delta) \big\rceil$
      \Comment{the aim, \cref{sec:method}}
      \State $p_j(k) \gets$ Eq.~\eqref{eq:pvec} at $\ask_k$, for every $j$
      \Comment{the kernel, run forward}
    \State $S_k \gets \sum_{j} p_j(k)\big(\widehat{Q}_j - \lambda z_j\big)
           + c_e/\sqrt{n_k + 1}$ \Comment{Eq.~\eqref{eq:acq}}
  \EndFor
  \State $k^\ast \gets \arg\max_{2 \le k \le K-1} S_k$
    \Comment{deterministic; the bonus already schedules variety}
  \State $\textit{mode} \gets \textsc{compress}$ if $\tau_{k^\ast}$ lies below
         the last parent of $P_{k^\ast}$, else $\textsc{expand}$
         \Comment{\cref{tab:breed}}
  \State $T \gets \textsc{llm}\big(\textsc{breed}[\textit{mode}],\, M,\,
         P_{k^\ast},\, \ask_{k^\ast}\big)$
  \State $n_{k^\ast} \gets n_{k^\ast} + 1$
    \Comment{the \emph{ask} is charged, measured or not}
  \If{$\textsc{measure}(T)$ succeeds}
    \State file $T$ in the niche of its \emph{written} size,
           $j \gets \mathrm{niche}(\landed(T))$
    \State $\Delta \gets$ increment of $\mathrm{MOQD}$, clipped to $[0,1]$
      \Comment{the reward}
    \State $\delta \gets (1-\eta)\,\delta + \eta\big(\ln \landed(T) - \ln \ask_{k^\ast}\big)$
      \Comment{LMS update}
    \State $\widehat{Q}_j \gets \widehat{Q}_j + \alpha\big(\Delta -
           \widehat{Q}_j\big)$
      \Comment{credited where it \emph{landed}}
    \State $\mathrm{hv}^{\mathrm{best}} \gets \max\big(\mathrm{hv}^{\mathrm{best}},
           \mathrm{hv}(T)\big)$;\ \ recompute $\lambda$ by Eq.~\eqref{eq:acq}
  \EndIf
\EndWhile
\State \Return the smallest-written suite among those of maximal $\mathrm{hv}$\Comment{\cref{def:task}}
\end{algorithmic}
\end{algorithm}

\section{Parent selection and context assembly}
\label{app:parents}
\label{ParentContext}
\label{sec:parents}

Given the selected niche $k^\ast$, \tesearch{} selects two parents from all
Pareto-front suites in the archive, rather than only from the target niche.
Specifically, one parent is taken from each side of the target size: on each
side, suites are ranked by their log-size distance to the target,
\[
\left|\ln \landed(T)-\ln \tau_{k^\ast}\right|,
\]
with hypervolume breaking ties, and the nearest suite is selected.
Using the global archive allows generation even when the target niche is empty,
while size proximity provides examples at a scale close to the requested suite.

The generation context consists of the module under test, execution feedback,
the two selected parents, and the requested size.
Execution feedback highlights uncovered lines and surviving mutants.
The parents are ordered from weaker to stronger, placing the stronger example
closest to the generation request.
Finally, the LLM is instructed to generate a suite of the requested size by
compressing or extending the last parent, according to whether the target lies
below or above it.
\Cref{app:prompts} provides the full prompt.

\paragraph{Parent selection.} Parents are selected by size proximity from the
two sides of the target: the Pareto-front suite nearest $\tau_{k^\ast}$ in log
size among those smaller than the target, and the nearest among those larger.
Because the grid is geometric, the niche index is log size up to a constant, so
each side is a one-dimensional nearest-neighbor lookup. The two parents
therefore always differ in size, and their relation in \cref{tab:annot} follows
from their hypervolumes: if the smaller parent has no lower hypervolume, it
Pareto-dominates the larger one; otherwise the larger one lexicographically
dominates it. When one side has no suite, the parents come from the
other side alone; in that edge case the call must compress or expand the suite
toward the target while raising its hypervolume. The pool is global rather than
niche-local for a structural reason: the acquisition frequently selects a niche
no suite has reached yet, and a draw restricted to that niche's own front would
leave exactly those targets unbreedable. The \emph{w/o nearest-parents} ablation
replaces the ranking with a uniform draw over the same pool and changes nothing
else --- the candidates are still front members, so nothing dominated is ever
shown, and the ordering rule below still applies.

\paragraph{Ordering and annotation.} The two parents and the target are
arranged ascending under the order of \cref{def:task} --- by hypervolume first,
and by smaller size only when hypervolumes are equal --- with the target always
last, immediately before the continuation, where recency weighs the nearest
example most \citep{zhao2021calibrate,liu2024lost,yang2024llmoptimizers}. With
one parent on each side of the target size, the arrangement takes one of two
mirror forms (\cref{tab:breed}). If the larger parent $L$ has the higher hypervolume, the order is
$S, L, T$: $L$ lexicographically dominates $S$, and the target $T$ must
Pareto-dominate $L$, compressing it to the target size at no lower hypervolume.
Otherwise the order is $L, S, T$: $S$ Pareto-dominates $L$, and $T$ must
lexicographically dominate $S$, expanding it to the target size at strictly
higher hypervolume. The asymmetry is deliberate: a smaller suite at equal
hypervolume is already an improvement under \cref{def:task}, and the only one
left once hypervolume saturates, whereas a larger suite is worth its extra tests
only if it raises hypervolume. The second parent carries a one-line note
comparing it with the first, in one of the two forms of \cref{tab:annot}; the
note states the measured change in both hypervolume and size, so it is always a
measured claim rather than an inferred narrative.

\begin{table}[h]
\centering
\caption{The comparison note attached to the second parent. Together with the
target, which always comes last, the ordered parents form a gradient along the
order on $(\mathrm{hv}, \text{size})$, so the strongest example sits nearest the
output position, where the model's recency bias weighs it most
\citep{zhao2021calibrate,liu2024lost,yang2024llmoptimizers}. Because the parents
lie on opposite sides of the target size and are ordered first, the later of the
two is never the weaker, so the relation reduces to two cases, and the note
reports both changes with their numbers.}
\label{tab:annot}
\footnotesize
\setlength{\tabcolsep}{5pt}
\renewcommand{\arraystretch}{1.2}
\begin{tabular}{l>{\raggedright\arraybackslash}p{3.6cm}>{\raggedright\arraybackslash}p{5.4cm}}
\toprule
\rowcolor{tblhead}
Type & Condition on (prev, curr) & Emitted note \\
\midrule
Pareto dominance & no lower hypervolume and smaller size &
\texttt{perfect improved version of prev (hypervolume grow 0.852->0.916, size
decline 24->13)} \\
\rowcolor{tblrow}
lexicographic dominance & higher hypervolume and larger size &
\texttt{improved version of prev (hypervolume grow \dots, size grow \dots)} \\
\bottomrule
\end{tabular}
\end{table}

\begin{table}[h]
\centering
\caption{What the target $T$ must achieve against the parent placed next to it.
$S$ and $L$ are the parents below and above the target size. The requirement is
asymmetric in hypervolume because the step toward $T$ changes size in opposite
directions.}
\label{tab:breed}
\footnotesize
\setlength{\tabcolsep}{4pt}
\renewcommand{\arraystretch}{1.2}
\begin{tabular}{>{\raggedright\arraybackslash}p{1.9cm}>{\raggedright\arraybackslash}p{1.8cm}>{\raggedright\arraybackslash}p{2.1cm}>{\raggedright\arraybackslash}p{6.4cm}}
\toprule
\rowcolor{tblhead}
Case (order) & $T$ against its neighbor & Hypervolume of $T$ & Emitted note \\
\midrule
compress ($S, L, T$) & smaller than $L$ &
$\mathrm{hv}(T) \ge \mathrm{hv}(L)$ &
\texttt{perfect improved version of prev (hypervolume grow or hold from
<hv(L)>, size decline <size(L)>-><n>)} \\
\rowcolor{tblrow}
expand ($L, S, T$) & larger than $S$ &
$\mathrm{hv}(T) > \mathrm{hv}(S)$ &
\texttt{improved version of prev (hypervolume grow from <hv(S)>, size grow
<size(S)>-><n>)} \\
\bottomrule
\end{tabular}
\end{table}

\section{Prompt construction}
\label{app:prompts}
\label{app:context}
\label{sec:context}

With the two parents chosen and ordered (\cref{app:parents}), three further
decisions turn them into a prompt: which of the two breeding framings the call
is given, what execution signal is written into the module source, and how much
of each piece is shown.

\paragraph{Feedback signal.} The module source shown in a breeding prompt is
annotated with what the \emph{strongest} parent still misses, written at the end
of the exact line it describes rather than as a separate list of line numbers:
\texttt{\# <-- NOT COVERED YET} on a line none of its passing tests execute, and
\texttt{\# <-- SURVIVING MUTANT: <op>} on a line where that mutation survives its
assertions. Failing tests inside an example suite are marked the same way, at the
end of the (possibly multi-line) \texttt{def} header, as
\texttt{\# <-- FAILING: <exception>}. Adjacency is the point: models align a mark
sitting next to the code far more reliably than a position given in a distant
block of text. Coverage only ever flags executable statement lines, so a trailing
comment is syntactically safe wherever a mark can land.

\paragraph{Prompts, verbatim.} Below are the three system prompts and the user
prompt skeleton, exactly as the implementation emits them; a bracketed
\texttt{[\dots]} marks text repeated verbatim from the prompt above it, and
\texttt{<\dots>} marks a slot filled at run time. Throughout this appendix,
\colorbox{promptsys}{\strut\small system prompts} are marked in blue and
\colorbox{promptusr}{\strut\small user prompts} in gray.

\begin{systempromptbox}[System Prompt --- \textsc{seed}]
\begin{lstlisting}[style=promptstyle]
You write ONE complete, self-contained Python test suite (pytest style) for the target module. Output ONLY the suite source - no prose, no fences. The suite must include its own fixture scaffolding at module level: import the SUT, bootstrap the runtime, stub collaborators, build inputs - then the tests. EVERY test must pass against the module as given. Compute expected values yourself by reading the module source. Prefer many small, strongly-asserting tests spread across ALL functions/classes of the module.
\end{lstlisting}
\end{systempromptbox}

\begin{systempromptbox}[System Prompt --- \textsc{breed}, compression mode]
\begin{lstlisting}[style=promptstyle]
You write ONE complete, self-contained Python test suite (pytest style) for the target module - a GLOBAL, whole-suite generation, not targeted at any one function or region. Output ONLY the suite source - no prose, no fences. The suite must include its own fixture scaffolding at module level: import the SUT, bootstrap the runtime, stub collaborators, build inputs - then the tests. EVERY test must pass against the module as given. Compute expected values yourself by reading the module source. You will be shown EXISTING versions of this suite, ordered from weakest to strongest by hypervolume (coverage x mutation-kill), the smaller one ranking stronger when hypervolumes are equal; the LAST one is the strongest shown. Each is labeled with its own size/coverage/mutation-kill numbers, and the second carries a short, factual note comparing it to the version immediately before it. Some tests in the versions you are shown FAIL against the module as given; each is marked `# <-- FAILING: <exception>` at the end of its `def` line, and the numbers in that version's header were measured on its PASSING tests only. Your task is CONSOLIDATION, not writing from scratch: the target size is BELOW the last version. Rewrite that version DOWN to the target size with a hypervolume at least as high as its own, and higher if you can: merge related assertions into fewer, denser tests; share setup; fold parametrizable cases together with pytest.mark.parametrize; drop only what is truly redundant. Do NOT drop that version's strong, exact-value assertions - they are what kills the mutants. Repair a failing test if its intent is worth keeping at the target size; otherwise drop it - it is contributing nothing to the numbers today.
\end{lstlisting}
\end{systempromptbox}

\begin{systempromptbox}[System Prompt --- \textsc{breed}, extension mode]
\begin{lstlisting}[style=promptstyle]
[...] Your task is EXTENSION, not writing from scratch: the target size is ABOVE the last version. Grow that version UP to the target size, and make the added tests raise its hypervolume: a larger suite whose hypervolume does not rise is worse than the version you started from. KEEP that version's tests and their strong, exact-value assertions (verbatim or lightly refactored), and ADD new, strongly-asserting tests that execute the `# <-- NOT COVERED YET` lines and assert exact values/branches around each `# <-- SURVIVING MUTANT` so the mutation would fail. Do not pad with redundant tests - every added test should close an annotated gap. Failing tests carry real testing intent - REPAIR them by reading the module source and fixing the expected value, rather than deleting them, unless the same behaviour is already checked elsewhere.
\end{lstlisting}
\end{systempromptbox}

The user prompt is the three pieces and the ask, in order: the module with the
residual signal marked inline, the count, then the ordered parents, strongest
last.
\begin{userpromptbox}[User Prompt --- \textsc{breed} (\texttt{<\dots>} are substituted at run time)]
\begin{lstlisting}[style=promptstyle]
[TARGET MODULE] <dotted.module.path>

[MODULE SOURCE - annotated with what the LAST (best) suite below still misses: `# <-- NOT COVERED YET` marks lines none of its tests execute; `# <-- SURVIVING MUTANT: <op>` marks lines where that mutation survives its assertions. Cover those lines, and assert exact values/branches around each surviving mutant so the mutation would fail]
<module source, in full, with inline residue marks>

[TARGET SIZE] exactly 18 test functions. This is a HARD REQUIREMENT: the suite MUST contain exactly 18 functions whose name starts with `test_`. Count them before you finish and adjust until the count is 18

[test_suite_v0 - size=24 green=22 pass_rate=0.917 coverage=0.947 mutation=0.900 loc=476]
(2 of its 24 tests FAIL - each marked inline below; the coverage/mutation above come from the 22 that pass)
<weaker parent, in full>

[test_suite_v1 - size=13 green=13 pass_rate=1.000 coverage=0.982 mutation=0.933 loc=273]
(perfect improved version of prev (hypervolume grow 0.852->0.916, size decline 24->13))
<stronger parent, in full>

Write test_suite_v2 now: the next, improved complete suite. Complete file, no prose, no fences.
\end{lstlisting}
\end{userpromptbox}

An arm whose niche the archive cannot supply parents for at all --- only possible
before the archive has any occupant --- falls back to the seed prompt with the
size instruction appended.

\section{Revision intents and examples}
\label{app:revisions}

The self-evolving experiment of \cref{sec:exp-selfevolve} revises eight modules
of mini-swe-agent~\citep{minisweagent}, chosen to cover the harness end to end:
the agent loop, the local execution environment, the two action parsers, the
cache-breakpoint policy, the message flattener, the model client and the config
merger (\cref{tab:signal} gives the role of each). For each module, GPT-5.4 is
shown the module source together with a revision intent and returns one
complete revised module. Intents are drawn from the fixed list in
\cref{tab:revision-taxonomy}, which covers the kinds of self-modification a
harness-evolving agent proposes in practice: a behavior-preserving intent asks
for the module to be improved without changing what it does, and a
behavior-changing intent asks for an edit that alters observable behavior.
\Cref{tab:revision-examples} shows representative revisions of each kind.

\begin{table}[t]
\caption{Revision intents. \textbf{Class} is what the intent asks for, not what
the revision turns out to do; a revision enters \cref{tab:signal} only if human
verification confirms that it realizes its intent (\cref{sec:exp-selfevolve}). Intents come in pairs: each behavior-changing intent is the failure
mode of the behavior-preserving intent above it, with the same goal and the same
shape of edit. $n$ counts revisions generated under each intent, pooled over the
eight modules.}
\label{tab:revision-taxonomy}
\centering
\footnotesize
\setlength{\tabcolsep}{5pt}
\renewcommand{\arraystretch}{1.05}
\begin{tabular}{l>{\raggedright\arraybackslash}p{0.58\textwidth}lc}
\toprule
\rowcolor{tblhead}
Family & Intent & Class & $n$ \\
\midrule
Performance & cache a function whose result depends only on its arguments & preserving & 16 \\
 & cache a function that reads changing state (clock, counter, file) & changing & 16 \\
 & list to set, where order and duplicates are unused & preserving & 16 \\
 & list to set, where order or duplicates are relied on & changing & 16 \\
 & batch independent calls, keeping results and their order & preserving & 16 \\
 & batch calls, changing which error surfaces and when & changing & 16 \\
 & return early where no side effect follows & preserving & 16 \\
 & return early, skipping a side effect (log, counter, cleanup) & changing & 16 \\
\midrule
Token & shorten the output, keeping every field a reader uses & preserving & 16 \\
 & shorten the output, changing a key or separator a reader parses & changing & 16 \\
 & truncate long output, keeping head and tail with a marker & preserving & 16 \\
 & truncate long output, keeping only the head & changing & 16 \\
 & stop emitting a field that no code reads & preserving & 16 \\
 & stop emitting a field read only on a rare path (e.g., errors) & changing & 16 \\
 & merge repeated outputs that are byte-identical & preserving & 16 \\
 & merge outputs that print alike but differ (e.g., exit code) & changing & 16 \\
\midrule
Robustness & retry a call that is safe to repeat & preserving & 16 \\
 & retry a call whose repetition has an extra effect & changing & 16 \\
 & add a timeout that reports expiry as a failure & preserving & 16 \\
 & add a timeout that returns partial output as complete & changing & 16 \\
 & narrow \texttt{except Exception} to every type the block raises & preserving & 16 \\
 & narrow \texttt{except Exception}, missing a type the block raises & changing & 16 \\
 & handle \texttt{None} where the value is legitimately optional & preserving & 16 \\
 & return \texttt{None} silently where the code should raise & changing & 16 \\
\midrule
Structure & extract a helper, keeping captured variables and defaults & preserving & 16 \\
 & extract a helper, changing a captured variable or default & changing & 16 \\
 & group parameters into a fresh object per call & preserving & 16 \\
 & group parameters into an object shared across calls & changing & 16 \\
 & merge two branches with identical bodies & preserving & 16 \\
 & merge two nearly identical branches, losing the difference & changing & 16 \\
 & add type hints and docstrings only & preserving & 16 \\
 & add type hints plus a small change to a conversion or default & changing & 16 \\
\midrule
Total & & & 512 \\
\bottomrule
\end{tabular}
\end{table}

\begin{table}[t]
\caption{Example divergent and equivalent revisions for six of the eight harness
modules, with the revision intent and content. \emph{eq} and \emph{div}
abbreviate equivalent and divergent.}
\label{tab:revision-examples}
\centering
\footnotesize
\setlength{\tabcolsep}{4pt}
\renewcommand{\arraystretch}{1.05}
\begin{tabular}{@{}l>{\raggedright\arraybackslash}p{0.25\textwidth}>{\raggedright\arraybackslash}p{0.39\textwidth}c@{}}
\toprule
\rowcolor{tblhead}
Module & Intent & What the revision did & Label \\
\midrule
\texttt{agents/default} & narrow \texttt{except Exception}, missing a type the block raises & narrows \texttt{except Exception} around the model call to \texttt{ValueError} and \texttt{TypeError}; a model raising \texttt{KeyError} or \texttt{OSError} now escapes the loop & div \\
\midrule
\texttt{environments/local} & return early, skipping a side effect (log, counter, cleanup) & checks for the submission sentinel only when the command failed, so a successful submit no longer ends the episode & div \\
 & extract a helper, keeping captured variables and defaults & moves the decoding of a failed command's partial output into a module-level helper, called at the same point on the same exception & eq \\
\midrule
\texttt{actions\_toolcall} & list to set, where order and duplicates are unused & tests for the required \texttt{command} argument with a one-element set's \texttt{issubset} instead of \texttt{in} & eq \\
\midrule
\texttt{actions\_text} & cache a function that reads changing state (clock, counter, file) & reads the clock once at import, so every observation carries the same stale timestamp & div \\
\midrule
\texttt{cache\_control} & stop emitting a field read only on a rare path (e.g., errors) & deletes both places that attach the ephemeral \texttt{cache\_control} marker, so no breakpoint is ever set & div \\
 & cache a function whose result depends only on its arguments & caches the content-text lookup behind a function that returns its argument, so a cache hit returns what a fresh call would & eq \\
\midrule
\texttt{serialize} & merge two branches with identical bodies & folds the merge-into-existing and fresh-copy branches into one call, passing \texttt{None} for a missing previous value, which the merge already skips & eq \\
\bottomrule
\end{tabular}
\end{table}

\clearpage
\section{Additional related work}
\label{app:additional_related}

\paragraph{Self-evolving agents in verticals.}
A general-purpose agent becomes useful in a vertical domain through knowledge
its weights do not hold, and with frozen weights (\cref{sec:related}) it evolves
by how that knowledge is streamed in, managed and retrieved. \emph{Streaming}
turns raw domain input, arriving continuously, into a structured form an agent
can consume: knowledge graphs constructed from unstructured threat
reports~\citep{cheng2025ctinexus}, hybrid system logs parsed by a user-adaptive
parser~\citep{xu2023hue}, and natural language translated into first-order
logic that symbolic reasoners can consume~\citep{putra2026nl2logic}.
\emph{Management} organizes and revises what has been accumulated: cyber threat
intelligence (CTI) corpora organized into agent-native scaffolds that agents
consume directly~\citep{cheng2026ctifoundry}, documentation optimized as
repository-specific playbooks for coding agents~\citep{cheng2026docsearch}, and
behavioral specifications of program entities refined recursively from the
failures of the code generated from them~\citep{cheng2026specforge}, where a
local improvement must be kept from causing a regression elsewhere.
\emph{Retrieval} decides what of it reaches the agent at decision time, as
benchmarked for retrieval-augmented LLMs over heterogeneous CTI
sources~\citep{cheng2026cticonnect}. Across the three stages, what lets a
vertical agent evolve rather than drift is a signal, grounded in its domain,
that decides which revisions of its knowledge to keep: knowledge graphs,
corpora and parsers are typically judged once, by extraction or answer
accuracy, and specifications and documentation by tests taken as given. This
paper turns to that signal itself. In software, where it is executable, we
search for the test suite that serves as it, strong enough to catch a revision
that silently changes behavior and small enough to run on every revision,
whether what is revised is documentation, a specification or, as in
\cref{sec:exp-selfevolve}, the agent's own harness.

\end{document}